# Cursive Multilingual Characters Recognition Based on Hard Geometric Features

**Amjad Rehman[1]**

[2]College of Business Administration Al Yamamah University Riyadh 11512 Saudi Arabia

Dr. Amjad Rehman is assistant professor in the MIS Department of Business Administration, Al Yamamah University. He earned PhD from Faculty of computing Universiti Teknologi Malaysia in 2010 as best university student. He has more than 100 publications indexed in Scopus and SCIE journals. His major research interests are health informatics, handwriting recognition and data mining.

**Majid Harouni[2]**

[1]Department of Computer Science, Dolatabad Branch Islamic Azad University, Isfahan, Iran.

Dr Harooni did his PhD from University Teknologi Malaysia with specialization in image processing and pattern recognition. Currently he is a assistant Prof. in Department of Computer Science, Dolatabad Branch Islamic Azad University, Isfahan, Iran.

**Tanzila Saba[3]**

[3]College of Computer and Information Sciences Prince Sultan University Riyadh 11586 Saudi Arabia

Dr. Tanzila Saba earned PhD in document information security and management from Faculty of Computing Universiti Teknologi Malaysia (UTM), Malaysia in 2012. She won best student award of 2012 in the faculty. Currently she is an Associate Prof. and IEEE senior member.

**ABSTRACT**

The cursive nature of multilingual characters segmentation and recognition of Arabic, Persian, Urdu languages have attracted researchers from academia and industry. However, despite several decades of research, still multilingual characters classification accuracy is not up to the mark. This paper presents an automated approach for multilingual characters segmentation and recognition. The proposed methodology explores character based on their geometric features. However, due to uncertainty and without dictionary support few characters are over-divided. To expand the productivity of the proposed methodology a BPN is prepared with countless division focuses for cursive multilingual characters. Prepared BPN separates off base portioned indicates effectively with rapid upgrade character acknowledgment precision. For reasonable examination, only benchmark dataset is utilized.

## 1. Introduction.

The unconstrained multilingual characters segmentation and recognition issues are still fresh [1-5]. The recognition rate is highly dependent on the segmentation accuracy. Therefore, innovative techniques are still desired to enhance characters segmentation and recognition accuracy on benchmark datasets [6-10]. Generally, characters recognition approaches could be classified into two categories; Online and offline characters recognition.

In most of the existing segmentation algorithms, human writing is evaluated empirically to derive rules [11-15]. Although the derived rules are satisfactory, there is no guarantee for their optimum results in all writing style. This is due to the human writing style that varies from person to person and even for the same person depending on different parameters. Hence, rules alone are not enough to segment and recognize multilingual characters. To overcome these limitations, researchers have employed artificial neural networks, hidden Markov models, and statistical classifiers to enhance segmentation accuracy [16-20]. Consequently, complex features are employed that raise issues of computational complexity and huge memory usage [21-25].

This research attempts to integrate, a rule-based segmentation approach and artificial neural network for the multilingual character recognition. The proposed approach performs character segmentation and recognition of isolated characters using BPN. Accordingly, BPN is trained with both correct and incorrect segmentation points for the words images of the benchmark database. This trained neural network is employed to verify the segmentation points marked by the proposed algorithm.

The rest of the paper is organized into four sections. Section 2 presents the proposed segmentation algorithm along with segmentation results. In section 3, neural-based experimentation is performed and results are discussed. Finally, conclusion and future works are drawn in section 4.

2. **Proposed Character Recognition Approach**

The proposed segmentation approach consists of three main phases: preprocessing, character segmentation algorithm and neural validation of the segmentation points. For the fair comparison, patterns are selected from the benchmark database. A few grayscale cursive handwritten samples for character segmentation, training, and testing of the BPN are shown in Figure 1.

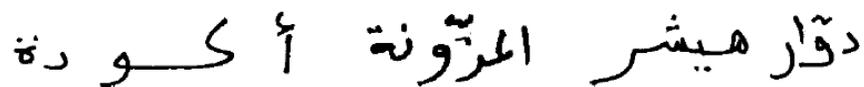

Fig 1. Samples of multilingual images

**2.1 Preprocessing**
Multilingual benchmark databases normally consist of characters/words/lines images in grayscale format (0-255). Therefore, images are a threshold to filter noise and to obtain a bi-level image (0-1) using the Otsu algorithm. There are many advantages of storing word images in this format. First, it is easy to deal with only two levels of colors and further processing will be fast, computationally less expensive, less storage demanding [26-30]. Additionally, in the proposed approach characters are segmented vertically therefore, to avoid the shadow of one character to the neighboring characters, characters slant is corrected. Finally, to work out with the geometrical features and to accommodate large variability of the handwriting stroke width, the thinning algorithm is applied. Preprocessing results are exhibited in Figure 2.

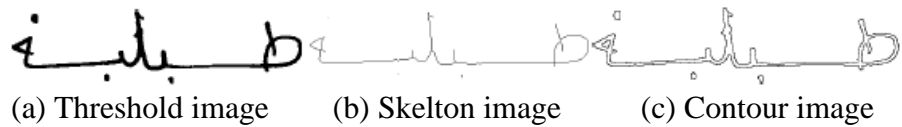

    (a) Threshold image    (b) Skelton image    (c) Contour image

Fig 2: Preprocessing

**2.2 Segmentation**

Multilingual alphabets are of three types [31-34]. First, typically consists of characters composed of the closed loop such as ھ , ö, ظ, the second category is of semi-loop such ک, ں and the third category of multilingual alphabets is similar to ligatures such as ⁄, ا[41]. Hence, character segmentation in Arabic script words is poised to overcome the following challenges:
  i. A ligature is a link between two characters in cursive handwriting. It is sometimes identical to a character or part of a character.
  ii. Characters in the word could be touching.
  iii. The ascenders and descenders may overlap with the neighboring characters.

The proposed character segmentation approach analyses the characters geometric features to identify ligatures and characters. Accordingly, the strategy consists of three main phases. In the first phase, closed characters are segmented. Secondly, boundaries of open characters are identified and finally, trained BPN is applied to enhance segmentation accuracy. The description of each phase is detailed below.

  Following preprocessing of cursive handwritten word images, ligatures are analyzed to segment closed characters. Accordingly, foreground pixels are counted in each column of the thinned word image as depicted in Figure 3. The columns are retained as candidate segmentation columns (CSC) for which counts are 0 or 1. This initial step is applied to locate boundaries of characters containing loop/semi-loop and to reduce the search domain. However, an over-segmentation problem is faced that emerges from two sources. First, the ligatures/ union of characters put forward many CSC. Second, open characters (without loop/ semi-loop) are also over-segmented due to the presence of within-letter-ligatures.

Therefore, additional features are needed to handle the problem of over-segmentation and to locate correct characters boundaries. The further process is detailed in the following steps.

In the first step, over-segmentation is eliminated in the ligatures. Accordingly, an average of those CSC is taken which is a distance less than a specified threshold to merge them into one segmentation column. Here, the threshold is the minimum horizontal distance between successive CSC that could not accommodate a character and is set to value 3 experimentally. This phase ensures valid segmentation of ligatures as well as reduces over-segmentation in the open characters such that an open character is over-segmented into maximum two segments.

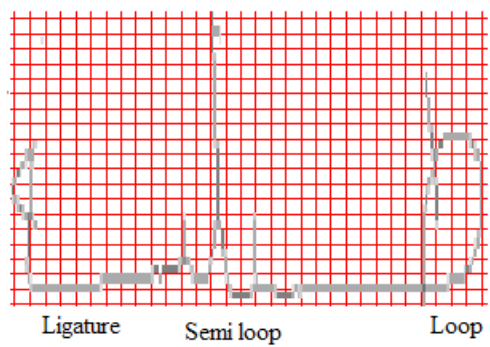

Fig 3: Foreground pixels counted to identify loops, semi-loops and ligatures

### 2.3 Characters recognition

To recognize multilingual segmented characters, a multilayer perceptron BPN with a back-propagation learning algorithm is employed [35-38]. However, it is worth to mention that over-segmentation is minimum and occurs for open characters only. Hence, it lessens the burden of the classifier employed and therefore, the processing speed is also increased. The pseudo code is presented below.

Step 1. Input an image from the multilingual database.
Step 2. Perform pre-processing.
Step 3. Calculate the sum of foreground pixels for each column. Save those columns as candidate segment column (CSC) for which sum is 0 or 1.
Step 4. By the previous step, we have more candidate segmentation columns than actually required. Accordingly, a threshold is selected empirically from candidate segment columns to come out with correct segment columns. A threshold is a constant value that is derived after a number of experiments to avoid over-segmentation.
Step 5. Finally, the back-propagation neural network is trained and applied for character segmentation's verification and to reject invalid segmentation point.

### 2.3.1 BPN Training and Testing

A simple program in MATLAB 7.0 is developed to detect coordinates of all segmentation points generated by the proposed segmentation technique for each pattern. These segmentation points are divided into correct and incorrect categories and are stored in a training file. Data is preprocessed prior to its use for BPN training as BPN takes uniform data [39-43].
To train BPN, the standard backpropagation algorithm is used. A number of experiments with different structures, weights, epochs, momentum and learning rate are performed to enable BPN to distinguish between correct and incorrect segmentation points. Finally, BPN trained with 1743 training patterns (segmentation points) taken from 1426 patterns. The optimal structure of BPN contains 261 to 321 inputs, 21 to 35 hidden units and one output (correct or incorrect confidence) with 315 epochs. Learning rate and momentum is set to 0.1 and 0.3 respectively. MATLAB 7.0 is used for implementation. Trained neural network operation is exhibited in Figure 4.

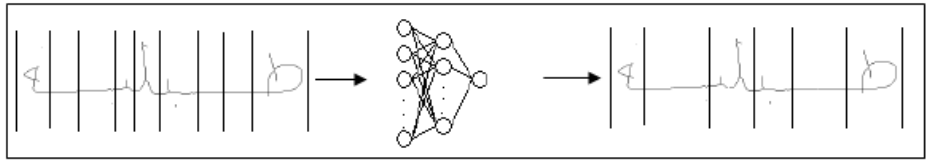

Fig 4. Incorrect Segmentation Points are rejected by Trained Neural Network.

Following training, 317-word images samples are selected from the dataset that is not used in the training. These new patterns are first segmented using the proposed algorithm. Thus obtained segmentation points are fed to the train BPN for their recognition into correct and incorrect categories. Finally, correct segmentation columns are retained while incorrect are extracted. Finally, character segmentation results for the test set are presented in Table 1.

TABLE 1: Segmentation Results (%)

| Correct segmentation rate. | 82.31 % |
|---|---|
| Miss-segmentation rate. | 2.95 % |
| Over-segmentation rate. | 3.24 % |
| Bad-segmentation rate | 4.73 % |

### 3. Experimental Results and Analysis

The characters geometric features analyzed with neural confidence achieved 82.31% segmentation accuracy for valid identification. However, two problems are found during the analysis of the results. Firstly, noisy characters, therefore some additional preprocessing (noise removal) is carried out before BPN training. Secondly, the presence of touched/

overlapped characters that is very hard to deal with. Ligatures could not be found, when two characters are tight together, and therefore, could not be segmented accurately [44-46]. Hence, overall segmentation accuracy is reduced.

Moreover, different researchers use different databases and report the results under some constraints. In literature, most of the researchers integrate segmentation approaches with some intelligent techniques such as neural networks, support vector machines, hidden Markov models to enhance accuracy. However, a brief comparison of achievement for the character recognition rate is presented. The following comparisons are selected as they are latest in the available literature and use the benchmark database [47-50].

Verma and Gader [57] obtain 76.52% accuracy rate using neuro-feature based approach on words taken from benchmark; however, the number of the word are not mentioned. Likewise, Blumenstein and Verma [58] claim 78.85% accuracy without mentioning a number of words. In the same way, Verma [59] claim 84.87% precision on 300 CEDAR benchmark. Similarly, Cheng et al., [60] acquire 95.27% segmentation rate from 317 CEDAR benchmark dataset. Finally, Cheng and Blumenstein [61] come out with 84.19% rate for 317 CEDAR benchmark dataset. Finally, it is stated that the proposed approach is reasonably fast due to minimum computational complexity. An average time taken for character recognition on the Pentium i7 processor is 0.057 sec/word. Segmentation results available in the literature are compared with the proposed approach in Table 2.

*TABLE 2: Comparison of accuracy rate*

| Author | Method | Accuracy rate (%) | Database used | **Comments** |
|---|---|---|---|---|
| Tappert et al [51] | Feature based + Rule based | 81.08 | NIST | number of words not mentioned |
| Han and Sethi [52] | Heuristic algorithm | 85.7% | Latin handwritten Words on 50 real mail envelopes | Only 50 mail envelopes are taken. |
| Lee et al [53] | BPN/MLP | 90 | Printed Latin alphanumeric characters | Printed alphanumerc characters used |
| Eastwood et al [54] | BPN/MLP | 75.9 | TMU database | 100,000 training pattern used |
| Blumenstein and Verma [55] | BPN + conventional method | 81.21 | 2568 words from | |
| *Yanikoglu and Sandon [56]* | *Linear Programming* | *97* | *750 words* | *No benchmark database used* |
| *Verma and Gader [57]* | *Feature based + BPN/MLP* | *91* | *IAM* | *words number not mentioned* |
| *Blumenstein and Verma [58]* | *Feature based+ BPN/MLP* | *78.85* | *CEDAR* | *words number not mentioned* |
| *Verma[59]* | *Feature based + BPN/MLP* | *84.87* | *CEDAR* | *300 words only* |
| *Cheng et al [60]* | *Feature based + BPN* | *95.27* | *TMU* | *317 words only* |
| *Cheng and Blumenstein [61]* | *Enhanced feature based+ BPN/MLP* | *84.19* | *CEDAR* | *317 words only* |
| **Proposed Approach** | *Geometric features and BPN/MLP* | *82.31* | *IFN/ENIT/TMU/ IAM/CEDAR* | *475 words.* |

4. **Conclusions**

This paper has exhibited that by just utilizing geometrical features of multilingual characters from Persian/Arabic/Urdu letters, their segmentation and recognition could be attained at high accuracy. These geometrical features include statistical and structure features are fed to trained BP/MLP neural network as a classifier. The evaluation of the system reveals that the proposed algorithm could support to find all meaningful geometric points of the different written styles of characters taken from different languages. Promising results achieved to recognize isolated Persian/Arabic/Urdu letters with an accuracy of 82.31% attained.

**REFERENCES**

[1] Harouni, M., Rahim,M.S.M., Al-Rodhaan,M., Saba, T., Rehman, A., Al-Dhelaan, A. "Online Persian/Arabic script classification without contextual information", The Imaging Science Journal, vol. 62(8), 2014, pp. 437-448, doi. 10.1179/1743131X14Y.0000000083.

[2] Muhsin; Z.F. Rehman, A.; Altameem, A.; Saba, A.; Uddin, M. "Improved quadtree image segmentation approach to region information". The imaging science journal, vol. 62(1), 2014, pp. 56-62, doi. http://dx.doi.org/10.1179/1743131X13Y.0000000063.

[3] Rehman, A. Kurniawan, F. Saba, T. "An automatic approach for line detection and removal without smash-up characters", The Imaging Science Journal, vol. 59(3), 2011, pp. 177-182, doi. 10.1179/136821910X12863758415649

[4] Saba, T. Rehman, A. Sulong, G. "Cursive script segmentation with neural confidence", International Journal of Innovative Computing and Information Control (IJICIC), vol. 7(7), 2011, pp. 1-10.

[5] Saba, T. Rehman, A. Altameem, A. Uddin, M. "Annotated comparisons of proposed preprocessing techniques for script recognition", Neural Computing and Applications, vol. 25(6), 2014, pp. 1337-1347, doi. 10.1007/s00521-014-1618-9

[6] Saba, T., Almazyad, A.S. Rehman, A. "Online versus offline Arabic script classification", Neural Computing and Applications,vol.27(7), 2016, pp 1797–1804, doi. 10.1007/s00521-015-2001-1.

[7] Saba, T. and Alqahtani, F.A. "Semantic analysis based forms information retrieval and classification", 3 D Research, vol. 4(4), 2013, doi. 10.1007/3DRes.03(2013)4.

[8] Saba, T., Rehman, A., and Sulong, G. "Improved statistical features for cursive character recognition" International Journal of Innovative Computing, Information and Control (IJICIC), vol. 7(9), 2011, pp. 5211-5224

[9] Fadhil, M.S. Alkawaz, M.H., Rehman, A., Saba, T. "Writers identification based on multiple windows features mining", 3D Research, vol. 7 (1), pp. 1-6, 2016, doi.10.1007/s13319-016-0087-6.

[10] Rehman, A. Kurniawan, F. Saba, T. (2011) "An automatic approach for line detection and removal without smash-up characters", The Imaging Science Journal, vol. 59(3), pp. 177-182, doi. 10.1179/136821910X12863758415649.

[11] Fadhil, M.S. Alkawaz, M.H., Rehman, A., Saba, T. (2016) "Writers identification based on multiple windows features mining", 3D Research, vol. 7 (1), pp. 1-6, doi.10.1007/s13319-016-0087-6.

[12] Rehman, A. and Saba, T. "Evaluation of artificial intelligent techniques to secure information in enterprises", Artificial Intelligence Review, vol. 42(4), 2014, pp. 1029-1044, doi. 10.1007/s10462-012-9372-9.

[13] Saba, T., Rehman, A., Al-Dhelaan, A., Al-Rodhaan, M. "Evaluation of current documents image denoising techniques: a comparative study". Applied Artificial Intelligence, vol.28 (9), 2014, pp. 879-887, doi. 10.1080/08839514.2014.954344

[14] S Joudaki, D Mohamad, T Saba, A Rehman, M Al-Rodhaan, A Al-Dhelaan (2014) Vision-Based Sign Language Classification: A Directional Review, IETE Technical Review, Vol.31 (5), 383-391, doi.10.1080/02564602.2014.961576.

[15] Nodehi, A. Sulong, G. Al-Rodhaan, M. Al-Dhelaan, A., Rehman, A. Saba, T. "Intelligent fuzzy approach for fast fractal image compression", EURASIP Journal on Advances in Signal Processing, 2014, doi. 10.1186/1687-6180-2014-112.

[16] Yousaf, K. Mehmood, Z. Saba, T. Rehman, A. Munshi,A.M. Alharbey, R. Rashid, M. (2019). Mobile-health applications for the efficient delivery of health care facility to people with dementia (PwD) and support to their carers: A survey. BioMed Research International,

vol. 2019, pp.1-26.

[17] Jadooki, S. Mohamad, D., Saba, T., Almazyad, A.S. Rehman, A. "Fused features mining for depth-based hand gesture recognition to classify blind human communication", Neural Computing and Applications, pp. 1-10, 2015, doi. 10.1007/s00521-016-2244-5

[18] Waheed, SR., Alkawaz, MH., Rehman, A., Almazyad,AS., Saba, T. "Multifocus watermarking approach based on discrete cosine transform", Microscopy Research and Technique, vol. 79 (5),2016,   pp. 431-437, doi. 10.1002/jemt.22646

[19] Rehman, A. and Saba, T. "Document skew estimation and correction: analysis of techniques, common problems and possible solutions", Applied Artificial Intelligence, vol. 25(9), 2011, pp. 769-787. doi. 10.1080/08839514.2011.607009

[20] Al-Turkistani, H. and Saba, T. "Collective intelligence for digital marketing", Journal of Business and Technovation, vol.3(3), 2015, pp. 194-203

[21] Rehman, D. Mohammad, G Sulong, T Saba (2009). Simple and effective techniques for core-region detection and slant correction in offline script recognition Proceedings of IEEE International Conference on Signal and Image Processing Applications (ICSIPA'09), pp. 15-20.

[22] Elsayed, H.A.G., Alharbi, A.N. AlNamlah, H. Saba, T. "Role of agile methodology in project management and leading management tools", Journal of Business and Technovation, vol.3(3), 2015, pp. 188-193

[23] Saba, T. "Pixel intensity based cumulative features for moving object tracking (MOT) in darkness", 3D Research, vol. 7(10), 2016, pp.1-6, doi. 10.1007/s13319-016-0089-4

[24] Neamah, K. Mohamad, D. Saba, T. Rehman, A. "Discriminative features mining for offline handwritten signature verification", 3D Research, vol. 5(2), 2014, doi. 10.1007/s13319-013-0002-3

[25] Arafat, S. and Saba, T. "Social media marketing vs social media commerce: a study on social media's effectiveness on increasing businesses' sales", Journal of Business and Technovation, vol. 4(3), 2016, pp. 134-147.

[26] Mundher, M. Muhamad, D. Rehman, A. Saba, T. Kausar, F. "Digital watermarking for images security using discrete slantlet transform", Applied Mathematics and Information Sciences,   vol. 8(6), 2014, pp. 2823-2830,   doi.10.12785/amis/080618.

[27] Lung, J.W.J Salam, M.S.H  Rehman, A. Rahim, M.S.M., Saba, T. "Fuzzy phoneme classification using multi-speaker vocal tract length normalization", IETE Technical Review, vol. 31 (2), 2014, pp. 128-136, doi. 10.1080/02564602.2014.892669

[28] Rehman, A. Kurniawan, F. Saba, T. "An automatic approach for line detection and removal without smash-up characters", The Imaging Science Journal, vol. 59(3), 2011, pp. 177-182, doi. 10.1179/136821910X12863758415649

[29] Rehman, A.  Alqahtani, S. Altameem, A. Saba, T. "Virtual machine security challenges:

case studies", International Journal of Machine Learning and Cybernetics vol. 5(5), 2014, pp. 729-742, doi. 10.1007/s13042-013-0166-4.

[30] Saba, T., Rehman, A., Sulong, G. "An intelligent approach to image denoising", Journal of Theoretical and Applied Information Technology, vol. 17 (2), 2010, pp. 32-36.

[31] MSM Rahim, SAM Isa, A Rehman, T Saba. "Evaluation of adaptive subdivision method on mobile device", 3D Research vol. 4(4),2013, pp. 1-10, doi:10.1007/3DRes.02(2013)4.

[32] Ahmad, AM., Sulong, G., Rehman, A., Alkawaz,MH., Saba, T. "Data hiding based on improved exploiting modification direction method and Huffman coding", Journal of Intelligent Systems, vol. 23 (4),2014, pp. 451-459, doi. 10.1515/jisys-2014-0007

[33] Rehman, A. and Saba, T. (2014). Neural network for document image preprocessing, Artificial Intelligence Review, Volume 42, Issue 2, pp 253-273, doi. 10.1007/s10462-012-9337-z.

[34] Saba, T. Rehman, A. Elarbi-Boudihir, M." Methods and strategies on off-line cursive touched characters segmentation: a directional review", Artificial Intelligence Review, vol. 42 (4), 2014, pp. 1047-1066. doi 10.1007/s10462-011-9271-5.

[35] Khan, A.R. and Mohammad, Z. (2008). A Simple Segmentation Approach for Unconstrained Cursive Handwritten Words in Conjunction with the Neural Network. International Journal of Image Processing, vol.2(3), pp. 29-35

[36] Saba,T. Almazyad, A.S., Rehman, A. (2015) Language independent rule based classification of printed & handwritten text, IEEE International Conference on Evolving and Adaptive Intelligent Systems (EAIS), pp. 1-4, doi. 10.1109/EAIS.2015.7368806.

[37] Ebrahim, A.Y., Kolivand, H., Rehman, A., Rahim, M.S.M. and Saba, T. (2018) 'Features selection for offline handwritten signature verification: state of the art', Int. J. Computational Vision and Robotics, vol. 8(6), pp.606–622

[38] Haji, M.S. Alkawaz, M.H. Rehman, A. Saba, T. (2019) Content-based image retrieval: a deep look at features prospectus, International Journal of Computational Vision and Robotics 9 (1), pp. 14-38.

[39] T.Saba and A. Rehman (2013). Effects of artificially intelligent tools on pattern recognition, International Journal of Machine Learning and Cybernetics, vol. 4(2), pp. 155-162. doi. 10.1007/s13042-012-0082-z.

[40] A Khalid, N Javaid, A Mateen, M Ilahi, T Saba, A Rehman (2019) Enhanced Time-of-Use Electricity Price Rate Using Game Theory, Electronics 8(1), 48.

[41] S Javaid, N Javaid, T Saba, Z Wadud, A Rehman, A Haseeb (2019) Intelligent Resource Allocation in Residential Buildings Using Consumer to Fog to Cloud Based Framework, Energies, vol.12(5), 815

[42] T Saba, SU Khan, N Islam, N Abbas, A Rehman, N Javaid, A Anjum (2019) Cloud‐based decision support system for the detection and classification of malignant cells in

breast cancer using breast cytology images, Microscopy research and technique, doi: 10.1002/jemt.23222.

[43] MA Khan, IU Lali, A Rehman, M Ishaq, M Sharif, T Saba, S Zahoor, et al (2019) Brain tumor detection and classification: A framework of marker‐based watershed algorithm and multilevel priority features selection, Microscopy research and technique, doi: 10.1002/jemt.23238

[44] A Sarwar, Z Mehmood, T Saba, KA Qazi, A Adnan, H Jamal (2019) A novel method for content-based image retrieval to improve the effectiveness of the bag-of-words model using a support vector machine, Journal of Information Science 45 (1), 117-135

[45] Sharif,U., Mehmood, Z. Mahmood, T., Javid, M.A. Rehman,A.  Saba, T. (2018) Scene analysis and search using local features and support vector machine for effective content-based image retrieval, Artificial Intelligence Review, 1-25

[46] Mehmood, Z. Gul, N. Altaf, M., Mahmood, T., Saba, T., Rehman, A., Mahmood, M.T. (2018) Scene search based on the adapted triangular regions and soft clustering to improve the effectiveness of the visual-bag-of-words model, EURASIP Journal on Image and Video Processing 2018 (1), 48

[47] Chaudhry, H., Rahim, MSM, Saba, T., Rehman, A. (2018)Crowd region detection in outdoor scenes using color spaces International Journal of Modeling, Simulation, and Scientific Computing, vol. 9(2), 1-15

[48] Mehmood, Z., Abbas, F., Mahmood, T. et al. Content-Based Image Retrieval Based on Visual Words Fusion Versus Features Fusion of Local and Global Features, Arab J Sci Eng (2018). https://doi.org/10.1007/s13369-018-3062-0

[49] S Jabeen, Z Mehmood, T Mahmood, T Saba, A Rehman, MT Mahmood (2018) An effective content-based image retrieval technique for image visuals representation based on the bag-of-visual-words model, PloS one 13 (4), e0194526

[50] T Mahmood, Z Mehmood, M Shah, T Saba (2018) A robust technique for copy-move forgery detection and localization in digital images via stationary wavelet and discrete cosine transform, Journal of Visual Communication and Image Representation, Vol.53, 202-214, https://doi.org/10.1016/j.jvcir.2018.03.015

[51] Tappert, C. C. Suen, C. Y. and Wakahara, T. (1990). The State of the Art in On-line Handwriting Recognition, IEEE Trans. on Pattern Analysis and Machine Intelligence, vol. 12(8), 787-793.

[52] Han, K., and Sethi, I., K. (1995). Off-line Cursive Handwriting Segmentation, Proceedings of the 3rd International Conference on Documents Analysis and Recognition, 894-897.

[53] Lee, H. and Verma, B. (2008a) A Novel Multiple Experts and Fusion Based Segmentation Algorithm for Cursive Handwriting Recognition. Proceedings of the International Joint Conference on Neural Networks (IJCNN'08), 2994-2999.

[54] Eastwood, B. Jennings, A. and Harvey, A. (1997). Neural Network Based Segmentation Handwritten Words. Proceedings of 6th International Conference on Image Processing and its Applications, Vol. 2, 750-755.

[55] Blumenstein, M. Liu, X.Y. and Verma. B. (2007). An Investigation of the Modified Direction Feature for Cursive Character Recognition, Pattern Recognition, vol. 40, 376–388.

[56] Yanikoglu, B., and Sandon, P., A. (1998). Segmentation of Online Cursive Handwriting using Linear Programming, Pattern Recognition, vol. 31, 1825-1833.

[57] Verma, B. Gader, P. Fusion of multiple handwritten word recognition techniques. Neural Networks for Signal Processing, Proceedings of the IEEE Signal Processing Society Workshop, 2, (2000), 926-934.

[58] Blumenstein, M., Verma, B. Analysis of segmentation performance on the CEDAR benchmark database, Proceedings of Sixth International Conference on Document Analysis and Recognition (2001), 1142.

[59] Verma, B. A contour character extraction approach in conjunction with a neural confidence fusion technique for the segmentation of handwriting recognition. Proceedings of the 9th International Conference on Neural Information Processing, 5,(2002), 18-22.

[60] Cheng, C.K., Liu, X.Y., Blumenstein, M. and Muthukkumarasamy, V. Enhancing neural confidence-based segmentation for cursive handwriting recognition. Proceeding of 5th International Conference on Simulated Evolution and Learning, (2004).

[61] Cheng, C.K., Blumenstein, M. Improving the segmentation of cursive handwritten words using ligature detection and neural validation, Proceedings of the 4th Asia Pacific International Symposium on Information Technology, (2005), 56-59.